\documentclass[letterpaper, 10 pt, conference]{ieeeconf}  %

\IEEEoverridecommandlockouts

\usepackage{cite}

\ifx\pdfoutput\undefined
\usepackage{graphicx}
\else
\usepackage[pdftex]{graphicx}
\fi

\usepackage{ifthen}
\usepackage[usenames,dvipsnames]{color}

\usepackage[font=small]{caption}

\usepackage{mathrsfs}     
\usepackage{amsmath}
\usepackage{amsfonts}
\usepackage{amssymb}
\usepackage{mathabx}
\usepackage{dsfont}
\usepackage{array}
\usepackage{wrapfig}
\usepackage{bigstrut}
\usepackage{multirow}
\usepackage{subcaption}
\usepackage{amssymb}
\usepackage{amsfonts}

\usepackage[ruled,linesnumbered,vlined]{algorithm2e}
\SetKwProg{Fn}{Function}{}{}
\SetArgSty{textrm}

\newtheorem{definition}{Definition}
\newtheorem{example}{Example}

\SetAlFnt{\small}

\begin{document}
%
\title{\LARGE \bf Traversing Environments Using Possibility Graphs with\\Multiple Action Types}


\author{Michael X. Grey$^{1}$ \and C. Karen Liu$^{2}$ \and Aaron D. Ames$^{3}$
\thanks{$^{1}$Michael X. Grey is a doctoral candidate in the School of Interactive Computing,
            Georgia Institute of Technology, Atlanta, GA 30332
            {\tt\small mxgrey@gatech.edu}}
\thanks{$^{2}$C. Karen Liu is an Associate Professor in the School of Interactive Computing,
            Georgia Institute of Technology, Atlanta, GA 30332
            {\tt\small karenliu@cc.gatech.edu}}
\thanks{$^{3}$Aaron D. Ames is an Associate Professor in the School of Mechanical Engineering \& the School of Electrical and Computer Engineering,
            Georgia Institute of Technology, Atlanta, GA 30332
            {\tt\small aaron.ames@me.gatech.edu}}
}

\maketitle

\begin{abstract}

Locomotion for legged robots poses considerable challenges when confronted by obstacles and adverse environments. Footstep planners are typically only designed for one mode of locomotion, but traversing unfavorable environments may require several forms of locomotion to be sequenced together, such as walking, crawling, and jumping. Multi-modal motion planners can be used to address some of these problems, but existing implementations tend to be time-consuming and are limited to quasi-static actions. This paper presents a motion planning method to traverse complex environments using multiple categories of continuous actions. To this end, this paper formulates and exploits the \emph{Possibility Graph}---which uses high-level approximations of constraint manifolds to rapidly explore the ``possibility'' of actions---to utilize lower-level single-action motion planners more effectively. We show that the Possibility Graph can quickly find routes through several different challenging environments which require various combinations of actions in order to traverse.

\end{abstract}

\IEEEpeerreviewmaketitle

\section{Introduction}

Modern motion planning methods have proven effective at navigating geometric constraint manifolds within high dimensional configurations spaces. This capability is critical for robots to exhibit autonomy in complex real-world environments, because geometric constraints are frequently used to determine the \emph{feasibility} of a physical action and hence are often used as ``feasibility constraints'' which must be satisfied or else the action is considered infeasible. Geometric constraints include requirements such as avoiding obstacles and placing end effectors in appropriate locations. Two common types of motion planners are Probabilistic Roadmaps (PRM) \cite{kavraki1998analysis} and Rapidly-exploring Random Tree (RRT) \cite{kuffner2000rrt}. Standard PRM is well-suited for exploring a single \emph{expansive} manifold, as defined in \cite{hsu1997path}. Constrained Bi-directional RRT (CBiRRT) \cite{cbirrt} can effectively handle constraint manifolds whose dimensionality is lower than the configuration space in which it exists.

Standard motion planning methods tend to struggle when a solution needs to traverse numerous discrete constraint manifolds. This occurs most often in hybrid dynamic systems where the ``mode'' of the system alters its constraint manifold. For example, standing on the left foot is a different mode from standing on both feet for a bipedal robot. The constraint manifolds of these two modes are different, and their intersection is narrow, resulting in a low (sometimes zero) probability of randomly moving from one manifold to the other. Hauser et al. introduced the Multi-modal PRM \cite{hauser2009multi} and Random-MMP \cite{hauser2011randomized} to address this problem. The primary bottleneck of this method is the combinatorial complexity of sampling and selecting modes, since each footstep taken by the robot represents a mode that must be explored. Additionally, existing methods for multi-modal planning are limited to quasi-static actions, which broadly eliminates their ability to utilize the dynamic capabilities of a robot system.

\begin{figure}
  \centering
  \begin{subfigure}[b]{\linewidth}
    \captionsetup{justification=centering}
    \centering
    \includegraphics[width=\linewidth]{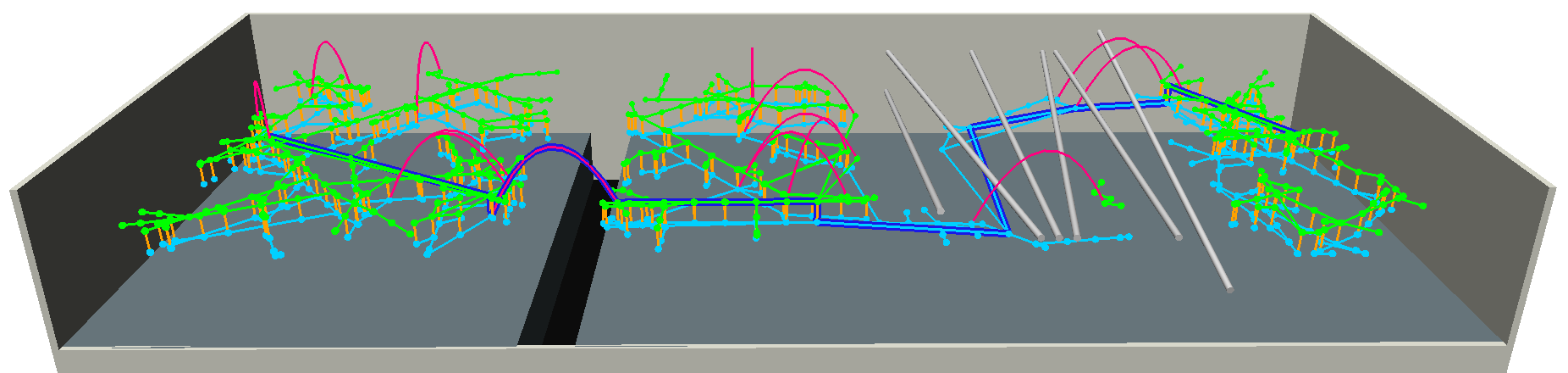}
    \caption{The graph explored the space of the hallway until a solution was found. Green edges are walking actions, light blue edges are crawling actions, and fuchsia arcs are jumping actions.}
    \label{fig:hallway_exploration}
  \end{subfigure}
  
  \begin{subfigure}[b]{\linewidth}
    \captionsetup{justification=centering}
    \centering
    \includegraphics[width=\linewidth]{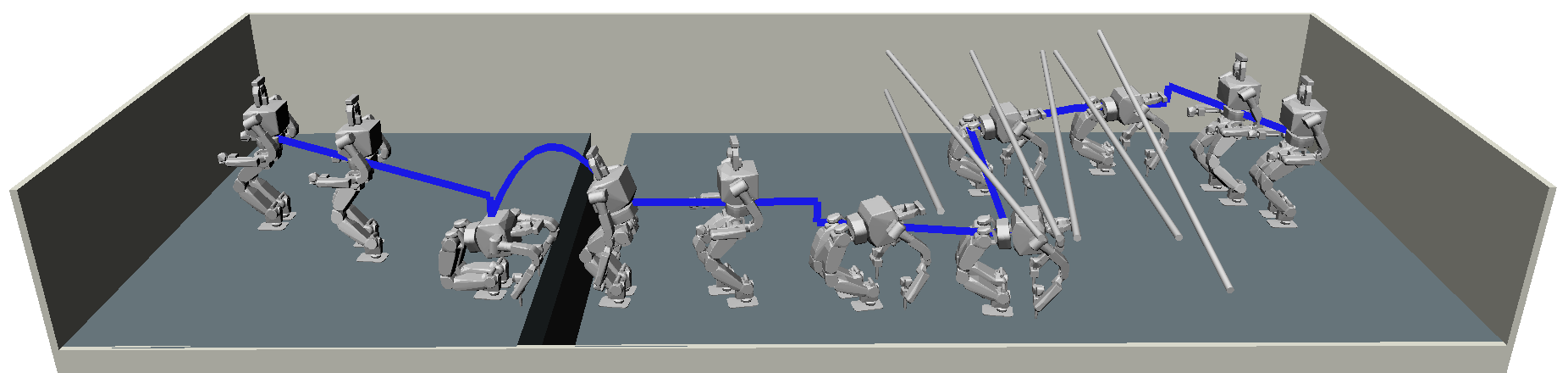}
    \caption{Snapshots showing the plan in action.}
    \label{fig:hallway_plan}
  \end{subfigure}
  
  \caption{\label{fig:hallway}The robot is tasked with traversing from the right side of a hallway to the left side. It must navigate underneath bars which are positioned at various angles, and then must jump across a gap.}
  
\end{figure}

In contrast to motion planning methods, standard footstep planners are able to rapidly generate footsteps and walking trajectories without spending time exploring the constraint manifolds of combinatorial modes the way Multi-modal PRM or Random-MMP do. They typically do this by approximating the problem of walking. In \cite{garimort2011humanoid, hornung2012anytime} this is done using a 2D representation of obstacles and in \cite{garimort2011humanoid, hornung2012anytime, candido2008improved, kuffner2001footstep, kuffner2003online, perrin2012fast, chestnutt2003planning} only a finite set of footstep parameters or action primitives are available to the planner. The two-stage method presented in \cite{pettre20032} uses a bounding cylinder to represent the collision geometry of the lower body. All of these estimations inherently limit the completeness of the methods. Moreover, these methods are all limited to a single category of action---bipedal walking---while being limited in terms of the dynamics of the motion.

The use of optimization methods in the motion planning domain has been growing \cite{zucker2013chomp, kuindersma2016optimization}, especially for walking motions. Nonlinear constrained optimization can elegantly handle the mixed modes and hybrid dynamics \cite{hereid20163d} required for walking and crawling. However, they tend to be tailored for generating single behaviors (e.g. a walking behavior that consists of a single- and double-support phase). This is insufficient for traversing a complex environment where a sequence of different types of behaviors is needed. Optimization methods also tend to be local, making them inappropriate for tackling problems that require a global search---especially for high dimensional humanoid robots.

The goal of this paper is to use a high-level motion planning method, named the \emph{Possibility Graph}, that can leverage the speed and efficiency of standard footstep planners with the completeness of randomized motion planning methods and the dynamics capabilities of optimization-based methods. The Possibility Graph is general enough to handle arbitrary categories of actions instead of being limited to only walking or stepping primitives. The role of the Possibility Graph is to quickly explore what actions might be possible throughout an environment. Different action types are compactly interlaced with each other within the graph, allowing a solution to utilize any action types in any order. Once a potential route is discovered, lower-level planning methods are used to confirm whether the route is truly feasible. This allows the lower-level (and computationally intensive) planners to focus their efforts on queries which are likely to achieve a solution. These queries can be performed in parallel, ensuring that the overall planning effort does not get bottlenecked by any single challenging step.

\section{Possibility Graphs for Multiple Action Types}\label{sec:posgraph}

Possibility Graphs were introduced in \cite{grey2017footstep} where they were used to guide Random-MMP searches through narrow passages in semi-unstructured environments. Possibility Graphs are effective at quickly exploring the environment at a high level to identify potential routes for the robot to follow. In this paper, we equip the Possibility Graph with multiple action types, allowing it to explore regions that would be unreachable with only one type of action. We augment the definition of the Possibility Graph from \cite{grey2017footstep} with a set of available action types to get Def. \ref{def:posgraph}:

\begin{definition}
\normalfont
\label{def:posgraph}A Possibility Graph is a tuple
\begin{equation}
PG = (\mathrm{actions}, \mathscr{E},\Gamma_{PG}=(V, E),Q_{Confirm})
\end{equation}
where,
\begin{itemize}
\item \texttt{actions} is the set of available action types,
\item $\mathscr{E}$ is the \emph{exploration space} for the graph (see Sec. \ref{sec:exploration_space}),
\item $\Gamma_{PG}$ is a graph consisting of vertices $V$ and edges $E$,
\item $V$ is a set of vertices which are elements of $\mathscr{E}$,
\item $E$ is a set of directed edges that transition between vertices using actions,
\item $Q_{Confirm}$ is a queue which manages confirmation jobs.\\
\end{itemize}
\end{definition}

\begin{figure}
  \centering
  \includegraphics[width=\linewidth]{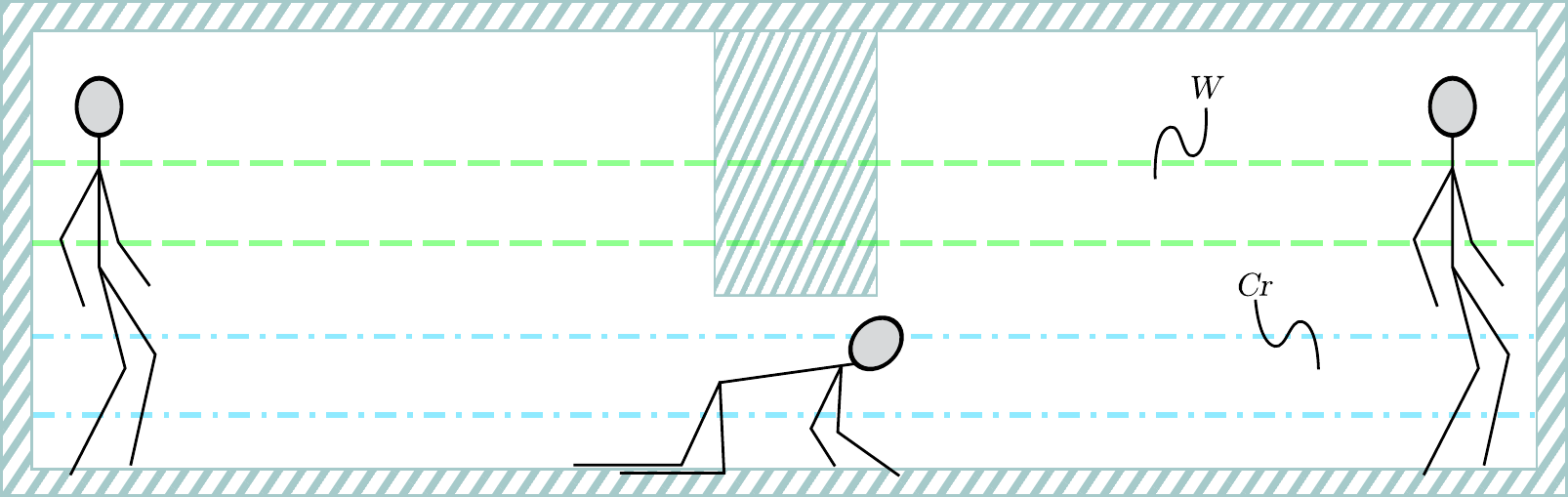}
  
  \vspace{2mm}
  \includegraphics[width=\linewidth]{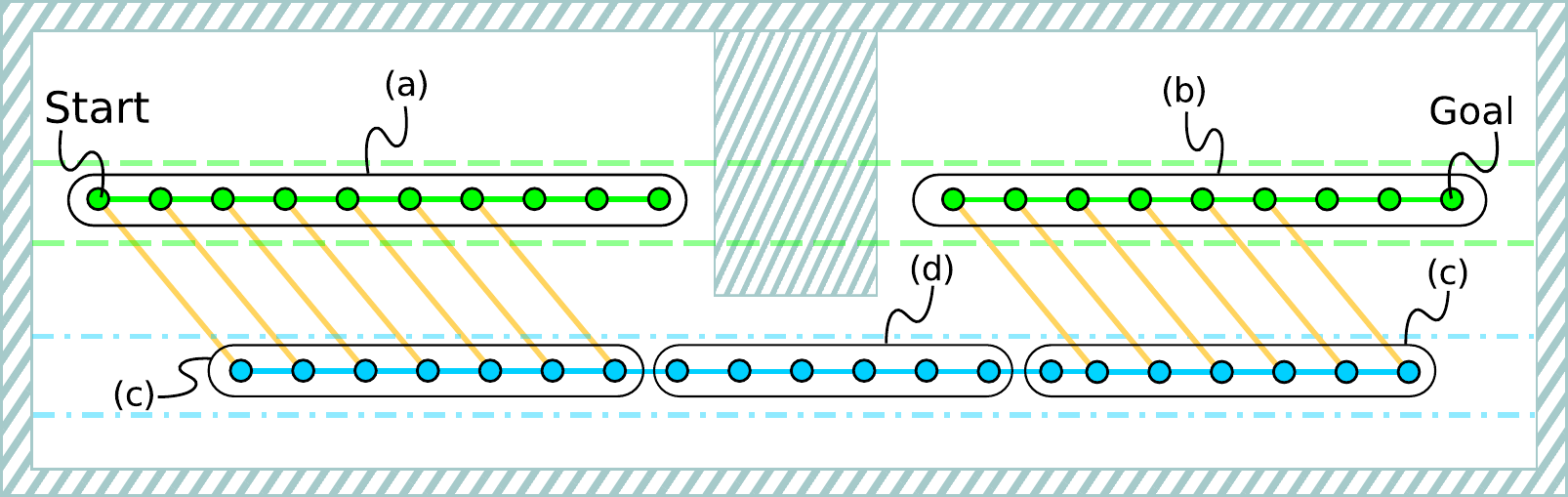}
  
  \caption{\label{fig:cartoon}Cartoon showing a simple 2D stick-figure example where the stick figure can walk or crawl forward. The graph's vertices represent the $(x,z)$ values of a point fixed to the stick figure's chest. The upper region, marked by $W$ in the top photo, is where walking is valid. The lower region, $Cr$, is where crawling is valid. (a) We extend from the start vertex towards a randomly sampled point in the center. [Alg. \ref{alg:holonomic_growth}, line \ref{alg:holonomic_growth:first_connect}] (b) We extend from the goal vertex towards the last vertex that was created in the previous step. [Alg. \ref{alg:holonomic_growth}, line \ref{alg:holonomic_growth:second_connect}] (c) For each new walking vertex, we create a crawling vertex and connect it to the walking vertex using a transition edge [Alg. \ref{alg:explore}, line \ref{alg:explore:transition}]. For some of the walking vertices, a transition into crawling is not viable due to obstacles. (d) We now extend the crawling subgraphs towards a point that was sampled near the center of the room, and the subgraphs manage to connect [Alg. \ref{alg:explore}, line \ref{alg:explore:growTowards}].}
\end{figure}

\begin{example}
In Fig. \ref{fig:cartoon} we show a Possibility Graph for a toy problem. The \texttt{actions} available to the stick figure are walking forward, crawling forward, and transitioning between walking and crawling. $\Gamma_{PG}$ can be seen in the lower panel: green edges and vertices belong to the walking action, blue edges and vertices belong to the crawling action, and orange edges are transitions where the robot is kneeling down or standing up.
\end{example}

\subsection{Sufficient vs. Necessary Conditions}

The defining feature of the Possibility Graph is the decomposition of an action's feasibility constraints into ``Sufficient Conditions'' and ``Necessary Conditions''. If an element of the graph satisfies the action's sufficient conditions, we can label it with ``Definitely Possible''. If an element violates the necessary conditions, then that element is ``Definitely Impossible'', and we exclude it from the graph. Otherwise, the element is labeled as ``Indeterminate'', because the necessary conditions are designed to be lax---they do not fully evaluate whether an element is feasible.

The sufficient conditions and necessary conditions can each define an explorable manifold as described in \cite{grey2017footstep}. As we build the Possibility Graph, the vertices of the graph will exist in the manifolds of these conditions instead of in the feasibility constraint manifold.

\begin{example}
In Fig. \ref{fig:cartoon} we show the necessary condition manifolds of walking and crawling. The stick figure must be standing in order to walk, so any vertices that are to be used for a walking action must lie between the dotted green lines. The stick figure must be on the ground in order to crawl, so only vertices that are between the dotted blue lines can be used for crawling.
\end{example}

\begin{algorithm}
\caption{\label{alg:explore}Finding a path using multiple action types}
\Fn{\texttt{FindPath}(start, goals, actions)}{
  $\Gamma_{PG}.V \gets \{\mathrm{start, goals}\}$\;
  Initialize each action graph with the start and goal vertices\;
  $Q_\mathrm{Confirm}$.\texttt{launchThreads}()\;
  $t \gets 0$\;
  \While{$t < t_{\mathrm{max}}$}{
    \For{$a$ in actions}{
      $\{V_\mathrm{new}, E_\mathrm{new}\} \gets a.\mathrm{\texttt{PerformTransitions}}()$\label{alg:explore:transition}\;
      $p_\mathrm{target} \gets$ \texttt{RandomSample}()\;
      $\{V_\mathrm{new}, E_\mathrm{new}\}$.\texttt{append}( $a$.\texttt{GrowTowards}($p_\mathrm{target}$) )\label{alg:explore:growTowards}\;
      \For{all $a_\mathrm{other}$ \textbf{not} $a$ in actions}{
        $a_\mathrm{other}.Q_\mathrm{Transition}.\mathrm{\texttt{insert}}(V_\mathrm{new})$\;
      }
      $\{\Gamma_{PG}.V, \Gamma_{PG}.E\}$.\texttt{append}($\{V_\mathrm{new}, E_\mathrm{new}\}$)\;
    }
    
    \For{$g$ in goals}{
      \If{\texttt{Connected}(start, g)}{
        $\Gamma_\mathrm{path} \gets$ \texttt{ShortestPath}(start, $g$)\;
        \If{\texttt{ConfirmPath}($\Gamma_{PG}$, $\Gamma_\mathrm{path}$, $Q_\mathrm{Confirm}$, actions)}{
          \Return $\Gamma_\mathrm{path}$\;
        }
      }
    }
    $t \gets$ \texttt{CurrentTime}()\;
  }
  \Return null\;
}
\end{algorithm}

\subsection{Exploration Space} \label{sec:exploration_space}

A key advantage of the Possibility Graph is it allows us to reduce the dimensionality of the problem's search space. The Possibility Graph is constructed based on the sufficient and necessary conditions of the actions that it is equipped with; therefore, it only needs to search over the parameters which are used as input arguments by the sufficient and necessary conditions of the available action types. We call this the ``Possibility Exploration Space'' and denote it as $\mathscr{E}$. This space is defined as the union of all the parameters that act as input arguments to the sufficient and necessary conditions of all available action types. Exploring the union of these parameters provides us with a unified space in which to explore the possibilities of multiple actions side-by-side.

In the toy problem of Fig. \ref{fig:cartoon}, the walking, crawling, and transitioning conditions can be defined as functions of the $(x,z)$ location of a point rigidly attached to the stick figure's chest. This makes $\mathscr{E} = \{x \in \mathbb{R}^2 \mid b_{\mathrm{lower}} \leq x \leq b_{\mathrm{upper}}\}$ where $b_{\mathrm{lower}}$ and $b_{\mathrm{upper}}$ are the lower and upper bounds of the environment. The parameter $z$ allows us to determine whether the stick figure may walk, crawl, or transition from a given vertex. $x$ lets us determine whether the stick figure would be in collision with the obstacle in the middle, and allows us to track whether we have progressed from the start to the goal. Later, when dealing with full 3D robots, we will use the SE(3) transform of a frame rigidly attached to the robot's pelvis for $\mathscr{E}$.

\subsection{Exploring Possibilities}

The purpose of the Possibility Graph is to find a feasible action sequence to get from a start point to at least one goal point. The procedure for finding solutions with the graph is described by Alg. \ref{alg:explore}, which is an augmented version of the \texttt{FindPath} algorithm of \cite{grey2017footstep}.

The important feature of this augmented multi-action version of the algorithm is the exploration of transitions between various action categories. Each time vertices are added for one action, the other actions will be queried to see if they can transition from it (Alg. \ref{alg:explore}, line \ref{alg:explore:transition}). This allows different actions to be interlaced with each other within the graph. Each action keeps track of its own exploration by storing a set of subgraphs, $\Gamma_a$, consisting of its own vertices and edges. At the same time, the Possibility Graph maintains the ``master'' graph, $\Gamma_{PG}$, which combines all the subgraphs of all the different action types. The algorithm is illustrated by a toy example in Fig. \ref{fig:cartoon}.

\begin{algorithm}
\caption{\label{alg:confirmation}Confirm a path}
\Fn{\texttt{ConfirmPath}($\Gamma_{PG}$, $\Gamma_\mathrm{path}$, $Q_\mathrm{Confirm}$, actions)}{
  pathConfirmed $\gets$ \textbf{true}\;
  \For{edge in $\Gamma_\mathrm{path}.E$}{
    edgeConfirmed $\gets$ \textbf{false}\;
    \For{$a$ in actions}{
      \If{$a.C_\mathrm{S}(\mathrm{edge})$}{
        edgeConfirmed $\gets$ \textbf{true}\;
      }
      \ElseIf{$a.C_\mathrm{N}(\mathrm{edge})$}{
        $\Gamma_{PG}$.\texttt{remove}(edge)\;
        $Q_\mathrm{Confirm}$.\texttt{insert}( $a$.\texttt{SpawnConfirmationJob}(edge) )\;
      }
    }
    \If{\textbf{not} edgeConfirmed}{
      pathConfirmed $\gets$ \textbf{false}\;
    }
  }
  \Return pathConfirmed\;
}
\end{algorithm}

\begin{algorithm}
\caption{\label{alg:transitions}Utilizing the Transition Queue}
\Fn{\texttt{Action::PerformTransitions}()}{
  $\{V_\mathrm{new}, E_\mathrm{new}\}  \gets$ \{new VertexQueue, new EdgeQueue\}\;
  $i \gets 0$\;
  \While{$i <$ MaxTransitionsPerCycle}{
    $v \gets \mathrm{\texttt{PopRandom}}(Q_\mathrm{Transitions})$\;
    $\{V_\mathrm{new}, E_\mathrm{new}\}$.\texttt{append}(\texttt{TransitionFrom}(v))\;
    $i \gets i+1$\;
  }
  $\Gamma_a$.\texttt{append}($\{V_\mathrm{new}, E_\mathrm{new}\}$)\;
  \Return $\{V_\mathrm{new}, E_\mathrm{new}\}$\;
}
\end{algorithm}

Over time, the Possibility Graph will consist of vertices and edges from various actions interlaced with each other. Some elements of the graph will satisfy the sufficient conditions of their respective actions, but some will only satisfy the necessary conditions. Once the graph contains a path from the start vertex to a goal vertex, we need to inspect the vertices and edges of that path to confirm whether all the path elements are truly feasible. This process is shown in the \texttt{ConfirmPath} function of Alg. \ref{alg:confirmation}. Each action type is responsible for spawning ``confirmation jobs'' which are low-level planning routines whose job is to verify whether or not an edge in the possibility graph is truly feasible. These routines are loaded into the Confirmation Queue, $Q_{Confirm}$. The Confirmation Queue will rotate between running each job to ensure that easy ones are finished promptly while difficult ones do not halt the overall confirmation progress. These jobs are executed on threads which run parallel to the graph expansion and each other. This allows the planner to search for alternative potential solutions when certain edges are difficult to confirm.


\section{Action Specifications}

For the Possibility Graph to explore actions, we need to fully define each action type. Table \ref{tab:action} lays out the implementation-dependent functions which must be engineered for each action type. The functions in that table enable the \texttt{GrowTowards} and \texttt{PerformTransitions} functions to work. \texttt{PerformTransitions} is described in Alg. \ref{alg:transitions}. It simply pulls vertices from other actions out of a queue and attempts to create transitions from those actions to itself. \texttt{GrowTowards} serves two primary roles: (1) expand the graph in new directions, and (2) connect together disjoint subgraphs. The nature of how an action grows will depend on what kind of action it is. In this paper we use three actions: walk, crawl, and standing long jump. These are divided into two categories: holonomic and nonholonomic.

\textbf{Holonomic actions} are expanded using Alg. \ref{alg:holonomic_growth}. When we describe an action as ``holonomic'' in this context, we mean that its sufficient/necessary condition manifold allows edges to branch into any direction at any time, just like the dynamics of a holonomic system. Even if the true physical dynamics of the action are nonholonomic, it can be treated as holonomic by the Possibility Graph if its necessary/sufficient condition manifold is simplified to behave holonomically within the exploration space, $\mathscr{E}$. Alg. \ref{alg:holonomic_growth} shows how the possibilities of holonomic actions are expanded. This is a modified version of the \texttt{GrowTowards} function from \cite{grey2017footstep} where the action type now keeps track of which vertices and edges it constructs by storing them in subgraphs.

\textbf{Nonholonomic actions} are expanded in a more complex way than holonomic actions, as shown in Alg. \ref{alg:nonholonomic_growth}. Nonholonomic actions generally cannot move directly towards a target, so they need to ``line themselves up'' first. We do this by identifying a launch point which is reachable from an existing point on the graph [Alg. \ref{alg:nonholonomic_growth}, line \ref{alg:launch}]. The launch point should be chosen such that it allows the action to land as close to the randomly generated target as possible, so long as the launch point is still reachable from the existing graph. Since nonholonomic actions are also generally direction-dependent, we do the reverse for goal-connected subgraphs [Alg. \ref{alg:nonholonomic_growth}, line \ref{alg:land}]: Pick a landing point which can connect to an existing goal-connected vertex such that it has a viable launch point coming from the direction of the target. Section \ref{sec:standing_long_jump} describes this for the jump action.

\begin{algorithm}
\caption{\label{alg:holonomic_growth}Growing the graph for a holonomic action}
\Fn{\texttt{Holonomic::GrowTowards}($p_\mathrm{target}$)}{
  $Q_\mathrm{closest} \gets$ new SortedVertexQueue\;
  \For{$g$ in $\Gamma_a$.SubGraphs}{
    $v \gets$ $g$.\texttt{FindClosestVertexTo}($p_\mathrm{target}$)\;
    $Q_\mathrm{closest}$.\texttt{insert}(dist($v$, $p_\mathrm{target}$), $v$)\;
  }
  $v_0 \gets Q_\mathrm{closest}$.\texttt{pop\_front}()\;
  $\{V_\mathrm{new}, E_\mathrm{new}\} \gets$ \texttt{Connect}($v_0$, $p_\mathrm{target}$)\label{alg:holonomic_growth:first_connect}\;
  $p_\mathrm{target} \gets V_\mathrm{new}$.\texttt{back}()\;
  \If{\texttt{UpstreamFromGoal}($v_0$)}{
    \While{\texttt{UpstreamFromGoal}($Q_\mathrm{closest}$.\texttt{front}()}{
      $Q_\mathrm{closest}$.\texttt{pop\_front}()\;
    }
  }
  $v_1 \gets Q_\mathrm{closest}$.\texttt{pop\_front}()\;
  $\{V_\mathrm{new}, E_\mathrm{new}\}$.\texttt{append}(\texttt{Connect}($v_1, p_\mathrm{target}$))\label{alg:holonomic_growth:second_connect}\;
  $\Gamma_a$.\texttt{append}($\{V_\mathrm{new}, E_\mathrm{new}\}$)\;
  \Return $\{V_\mathrm{new}, E_\mathrm{new}\}$\;
}
\vspace{1mm}
\Fn{\texttt{HolonomicAction::Connect}($v_\mathrm{start}$, $p_\mathrm{target}$)}{
  $\{V_\mathrm{new}, E_\mathrm{new}\}  \gets$ \{new VertexQueue, new EdgeQueue\}\;
  $v_\mathrm{last} \gets v_\mathrm{start}$\;
  $v \gets$ \texttt{ExtendTowards}($v_\mathrm{start}, p_\mathrm{target}$)\;
  $v_p \gets$ \texttt{Project}(v)\;
  \While{$C_\mathrm{N}(v_p)$ \textbf{and} $v \neq p_\mathrm{target}$}{
    edge $\gets$ Edge($v_\mathrm{last}, v_p$)\;
    \If{\textbf{not} $C_\mathrm{N}$(edge)}{
      \textbf{break}\;
    }
    $\{V_\mathrm{new}, E_\mathrm{new}\}$.\texttt{append}($\{v_p, \mathrm{edge}\}$)\;
    $v_\mathrm{last} \gets v_p$\;
    $v \gets$ \texttt{ExtendTowards}($v, p_\mathrm{target}$)\;
    $v_p \gets$ \texttt{Project}($v$)\;
  }
  \Return $\{V_\mathrm{new}, E_\mathrm{new}\}$\;
}
\end{algorithm}

\begin{algorithm}
\caption{\label{alg:nonholonomic_growth}Growing the graph for a nonholonomic action}
\Fn{\texttt{Nonholonomic::GrowTowards}($p_\mathrm{target}$)}{
  $\{V_\mathrm{new}, E_\mathrm{new}\}  \gets$ \{new VertexQueue, new EdgeQueue\}\;
  $Q_\mathrm{closest} \gets$ new SortedVertexQueue\;
  \For{$v$ in $\Gamma_a.V$}{
    $Q_\mathrm{closest}$.\texttt{insert}(dist($v$, $p_\mathrm{target}$), $v$)\;
  }
  Useful $\gets$ new BooleanArray($\Gamma_a.V$.size(), true)\;
  \For{$v$ in $Q_\mathrm{closest}$}{
    \lIf{\textbf{not} Useful[$v$]}{ continue }
    \If{\textbf{not} \texttt{UpstreamFromGoal}($v$)}{
      $v_\mathrm{launch} \gets$ \texttt{FindLaunchPoint}($v$, $p_\mathrm{target}$)\label{alg:launch}\;
      $v_\mathrm{landing} \gets$ \texttt{ExtendTowards}($v_\mathrm{launch}, p_\mathrm{target}$)\;
      edge $\gets$ Edge($v_\mathrm{launch}$, $v_\mathrm{landing}$)\;
      \If{$C_\mathrm{N}$(edge)}{
        $\{V_\mathrm{new}, E_\mathrm{new}\}.$\texttt{append}($\{v_\mathrm{launch}, v_\mathrm{landing}, \mathrm{edge}\}$)\;
        \texttt{SetUpstreamVerticesToFalse}($v$, Useful)\;
      }
    }
    \If{\textbf{not} \texttt{DownstreamFromStart}($v$)}{
      $v_\mathrm{landing} \gets$ \texttt{FindLandingPoint}($v$, $p_\mathrm{target}$)\label{alg:land}\;
      $v_\mathrm{launch} \gets$ \texttt{ReverseExtend}($v_\mathrm{landing}$, $p_\mathrm{target}$)\;
      edge $\gets$ Edge($v_\mathrm{launch}$, $v_\mathrm{landing}$)\;
      \If{$C_\mathrm{N}$(edge)}{
        $\{V_\mathrm{new}, E_\mathrm{new}\}$.\texttt{append}$\{v_\mathrm{launch}, v_\mathrm{landing}, \mathrm{edge}\}$\;
        \texttt{SetDownstreamVerticesToFalse}($v$, Useful)\;
      }
    }
  }
  $\Gamma_a$.\texttt{append}($\{V_\mathrm{new}, E_\mathrm{new}\}$)\;
  \Return $\{V_\mathrm{new}, E_\mathrm{new}\}$\;
}
\end{algorithm}

\begin{table}[h]
\caption{\label{tab:action}Definition of an Action}
\begin{tabular}{lll}
\multicolumn{2}{l}{\textbf{All action types}}\vspace{1mm}\\
\texttt{ExtendTowards}($v_0$, $v_1$):   & Create a vertex by moving towards $v_1$\\
                                        & from $v_0$ via this action.\vspace{1mm}\\
$C_N(x)$:                               & Return \textbf{true} if $x$ meets the action's\\
                                        & necessary conditions, otherwise return\\
                                        & \textbf{false}. $x$ may be a vertex or an edge.\vspace{1mm}\\
$C_S(x)$:                               & Return \textbf{true} if $x$ meets the action's\\
                                        & sufficient conditions, otherwise return\\
                                        & \textbf{false}. $x$ may be a vertex or an edge.\vspace{1mm}\\
\texttt{TransitionFrom}($v$):           & Attempt to return a path that goes\\
                                        & from $v$ into the necessary condition\\
                                        & manifold of this action.\vspace{1mm}\\
\texttt{SpawnConfirmationJob}($e$):     & Return a routine (called a confirmation\\
                                        & job) which can examine edge $e$ to\\
                                        & ascertain whether it is truly feasible.\vspace{3mm}\\
\multicolumn{2}{l}{\textbf{Holonomic action types}}\vspace{1mm}\\
\texttt{Project}($v$):                  & Attempt to return a point on the\\
                                        & necessary condition manifold which is\\
                                        & close to $v$.\vspace{3mm}\\
\multicolumn{2}{l}{\textbf{Nonholonomic action types}}\vspace{1mm}\\
\texttt{ReverseExtend}($v_0$, $v_1$):   & Create a vertex which can arrive at $v_0$\\
                                        & from the direction of $v_1$ via this action.\vspace{1mm}\\
\texttt{FindLaunchPoint}($v$, $v_1$):   & Return a point, $v_0$, close to $v$ which\\
                                        & can be used in a call to\\
                                        & \texttt{ExtendTowards}($v_0$, $v_1$)\vspace{1mm}\\
\texttt{FindLandingPoint}($v$, $v_1$):  & Return a point, $v_0$, close to $v$ which\\
                                        & can be used in a call to\\
                                        & \texttt{ReverseExtend}($v_0$, $v_1$)\\
\end{tabular}
\end{table}

\subsection{Walk and Crawl}

\begin{figure}
  \centering
  \begin{subfigure}[b]{0.29\linewidth}
    \captionsetup{justification=centering}
    \centering
    \includegraphics[width=\linewidth]{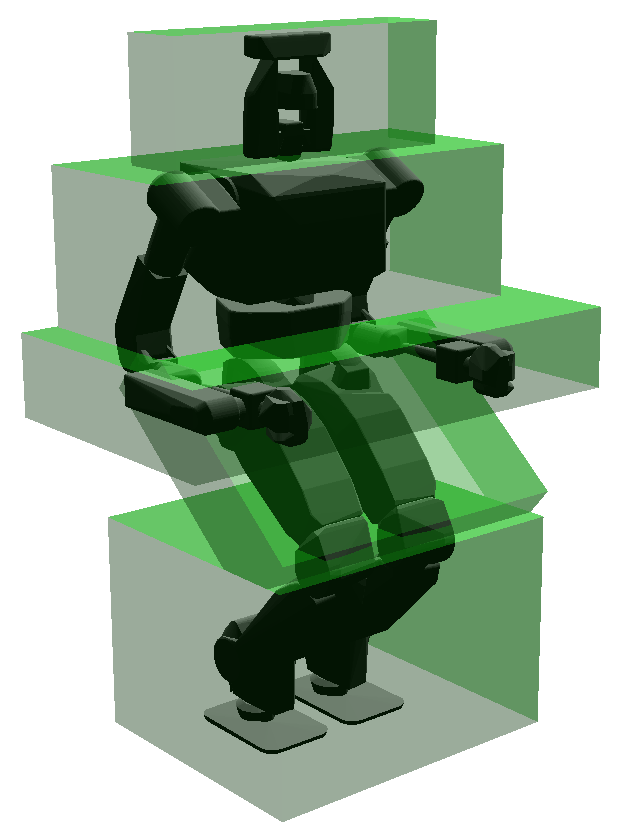}
    \caption{}
    \label{fig:walk_sweep}
  \end{subfigure}
  \begin{subfigure}[b]{0.39\linewidth}
    \captionsetup{justification=centering}
    \centering
    \includegraphics[width=\linewidth]{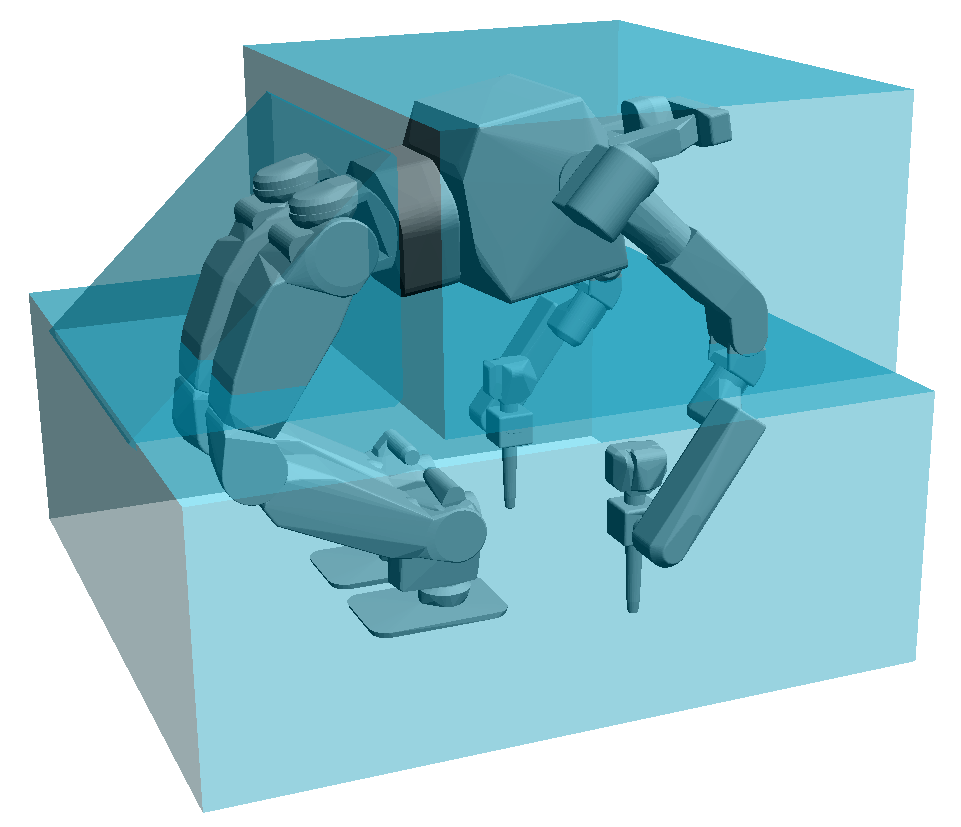}
    \caption{}
    \label{fig:crawl_sweep}
  \end{subfigure}
  \begin{subfigure}[b]{0.29\linewidth}
    \captionsetup{justification=centering}
    \centering
    \includegraphics[width=\linewidth]{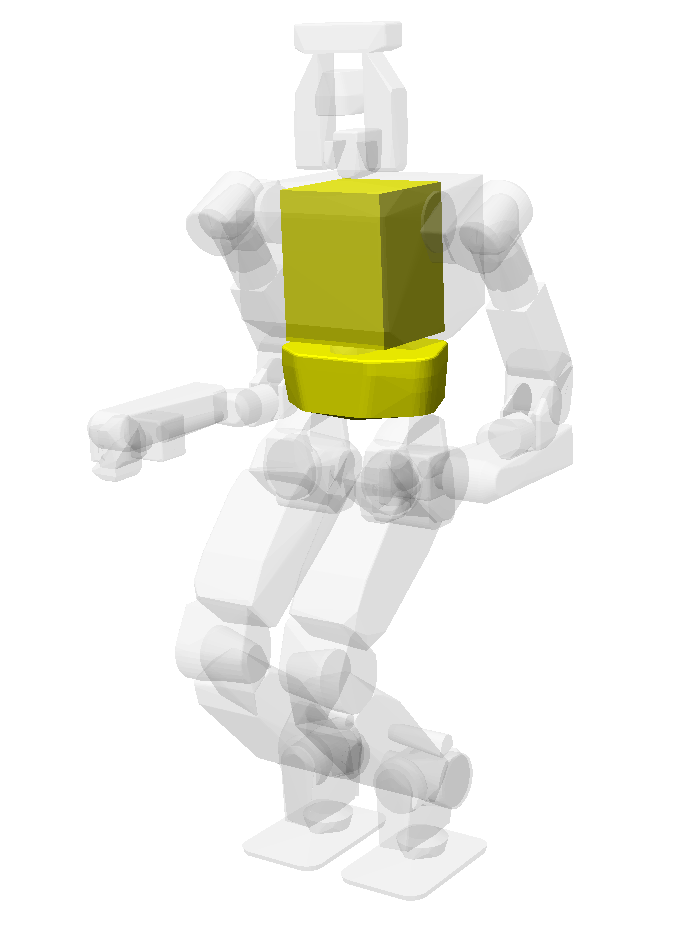}
    \caption{}
    \label{fig:minimal_sweep}
  \end{subfigure}
  
  \caption{\label{fig:sweeps}The nominal configurations used for (a) walking and (b) crawling, with their swept geometries surrounding them. (c) shows the collision geometry for the necessary conditions in yellow.}
\end{figure}

The walking and crawling actions are formulated very similarly to each other. These are the same conditions used for walking by \cite{grey2017footstep}. Sufficient conditions for walking and crawling are holonomic, and include these simplifications:

\begin{enumerate}
  \item The swept geometries in Fig. \ref{fig:sweeps} must not be in collision with the environment.
  \item Each point that defines the robot's support polygon must be touching flat ground when the robot is in a ``nominal'' walk/crawl configuration. The nominal configurations can be seen in Fig. \ref{fig:sweeps}.
  \item The root must be in the ``nominal'' orientation of the action (upright for walking and pitched backwards $80^\circ$ for crawling).
\end{enumerate}

The necessary condition is easier to satisfy: We use only the collision geometry seen in Fig. \ref{fig:minimal_sweep}, because all other bodies depend on joint parameters which are not included in $\mathscr{E}$.


\begin{figure}
  \centering

  \includegraphics[width=0.15\linewidth]{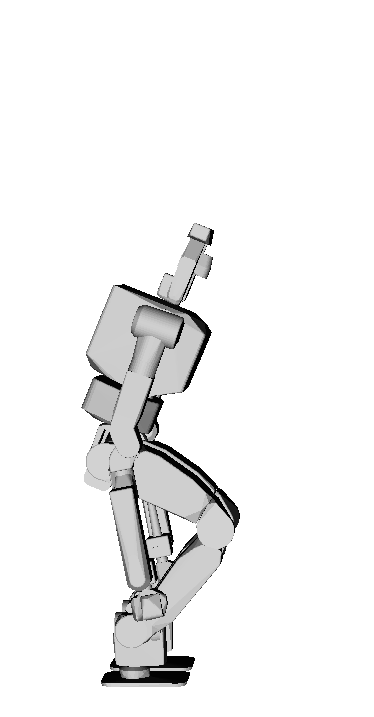}
  \includegraphics[width=0.15\linewidth]{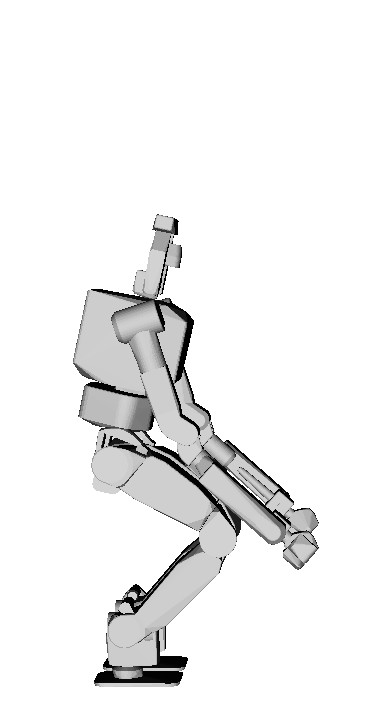}
  \includegraphics[width=0.15\linewidth]{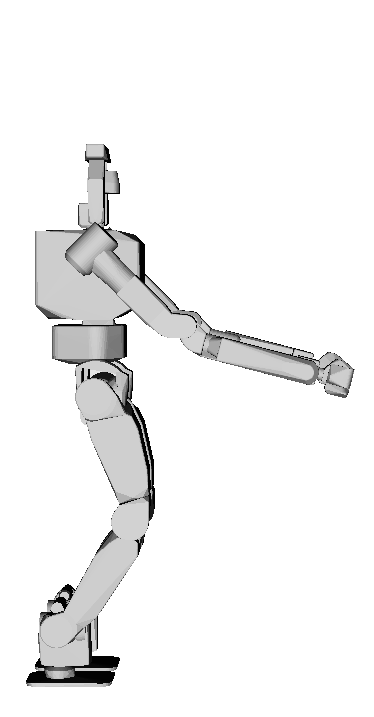}
  \includegraphics[width=0.15\linewidth]{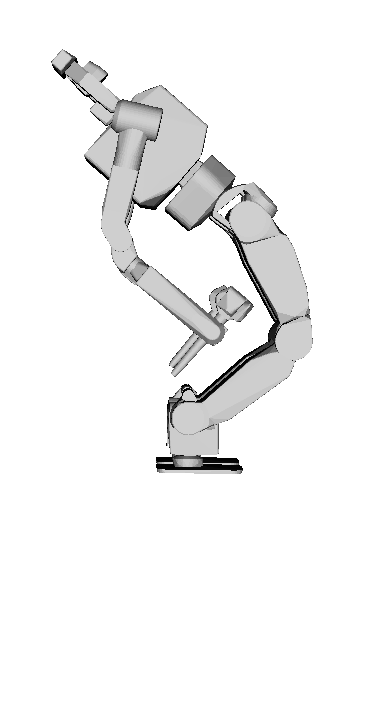}
  \includegraphics[width=0.15\linewidth]{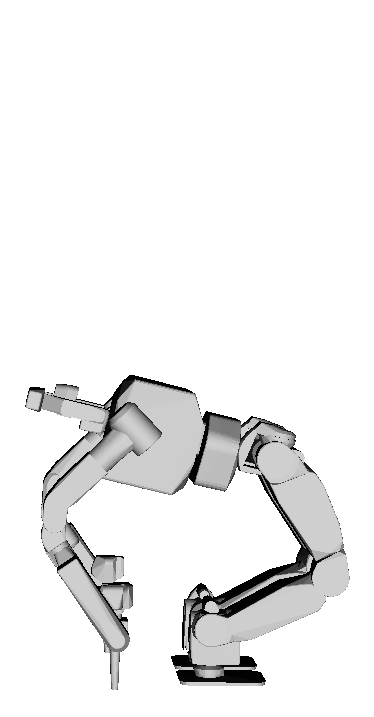}
  \includegraphics[width=0.15\linewidth]{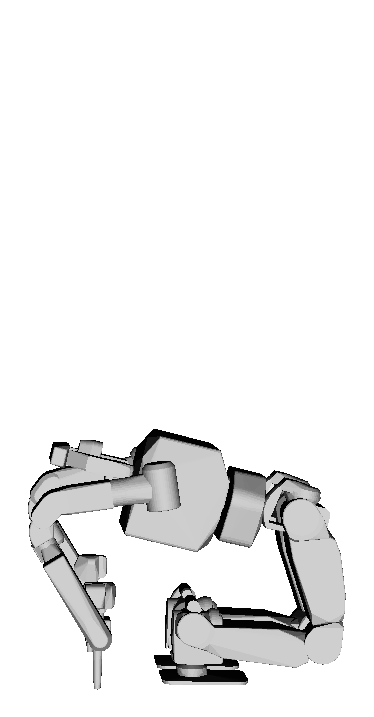}
  
  \caption{\label{fig:jump}An example standing long jump trajectory. The robot begins from a standing configuration, swings its arms, and jumps forward. It plans out its angular momentum so that it is able to land in a crawling configuration. After hitting the ground, it absorbs some of the impact by letting its joints behave elastically.}
\end{figure}

\subsection{Standing Long Jump}
\label{sec:standing_long_jump}
A standing long jump is a forward jump which begins from standing in place and launches forward without taking any steps. Figure \ref{fig:jump} shows an example of a jumping trajectory. We use a standing long jump in this paper for simplicity; future work will include long jumps that take running starts, which can achieve considerably greater range. We provide necessary conditions for the standing long jump but not sufficient conditions. The necessary condition manifold is nonholonomic, and contains the following:

\begin{enumerate}
  \item The jump must begin from a valid walk vertex.
  \item The jump must finish at a valid crawl vertex.
  \item There must be at least one collision-free parabola through $\mathscr{E}$ from the beginning vertex to the finishing vertex. The parabola must follow a feasible jump arc according to the physical limitations of the robot.
\end{enumerate}

The \texttt{TransitionFrom} function for the jump action is trivial, because it always begins from valid walking configurations and ends in valid crawling configurations, therefore the transition function does nothing. The \texttt{ExtendTowards}($v_0, v_1$) function performs a forward jump from $v_0$ to $v_1$. If $v_1$ is too far to reach from $v_0$, then it performs the furthest allowable jump. The \texttt{ReverseExtend}($v_0, v_1$) function instead performs a jump which lands at $v_0$ and begins as close to $v_1$ as the robot's physical limitations allow. The \texttt{FindLaunchingPoint}($v, v_1$) function returns a point, $v_0$, whose translation is the same as $v$ but whose orientation has the robot facing $v_1$; this allows the \texttt{ExtendTowards}($v_0, v_1$) function to jump towards $v_1$. Conversely, the \texttt{FindLandingPoint}($v, v_1$) function returns a point, $v_0$, whose translation is the same as $v$ but which is facing \emph{away} from $v_1$; this allows the robot to jump towards $v$ from the direction of $v_1$ using \texttt{ReverseExtend}($v_0, v_1$).

The \texttt{SpawnConfirmationJob} function of the jump action is a basic collocation optimization on a boundary value problem. The boundary value constraints are (1) zero initial velocity, and (2) a take-off configuration and velocity which will allow the robot to reach its jump target. The objective function of the optimization problem attempts to minimize the accelerations during take-off. While generating the trajectory, we also check that the joint and contact forces required to achieve the trajectory are physically feasible. Trajectories which fail this test are discarded. Once the jump is generated, we can check for collisions along its trajectory. If the jump was successfully generated (i.e. the jumping motion is physically feasible) and is collision-free, then its ``possibility'' status is changed from ``indeterminate'' to ``possible'', and it can be used in a final solution.

\section{Experiments}\label{sec:experiments}
\label{sec:experimental}

We run performance tests on three scenarios (one of which has three versions). Each performance test is the result of 50 trials. The Possibility Graph is a randomized planner, so the time required for the same trial can vary between runs. We put a 60 second time limit on the planner; if a solution is not found within 60 seconds, we consider it a failed run.

\begin{figure}
  \centering
  \begin{subfigure}[b]{0.98\linewidth}
    \captionsetup{justification=centering}
    \centering
    \includegraphics[width=\linewidth]{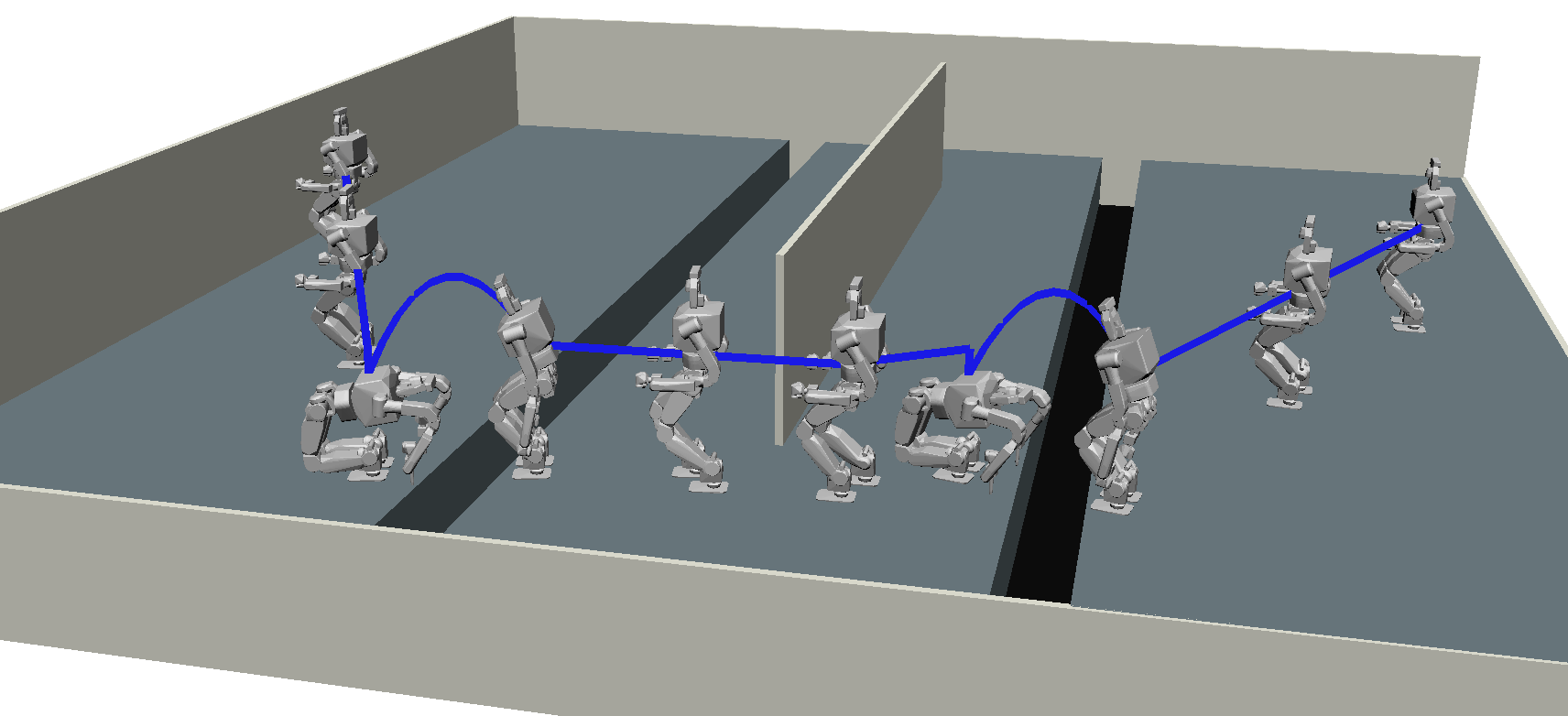}
  \end{subfigure}
  
  \begin{subfigure}[b]{0.98\linewidth}
    \captionsetup{justification=centering}
    \centering
    \includegraphics[width=\linewidth]{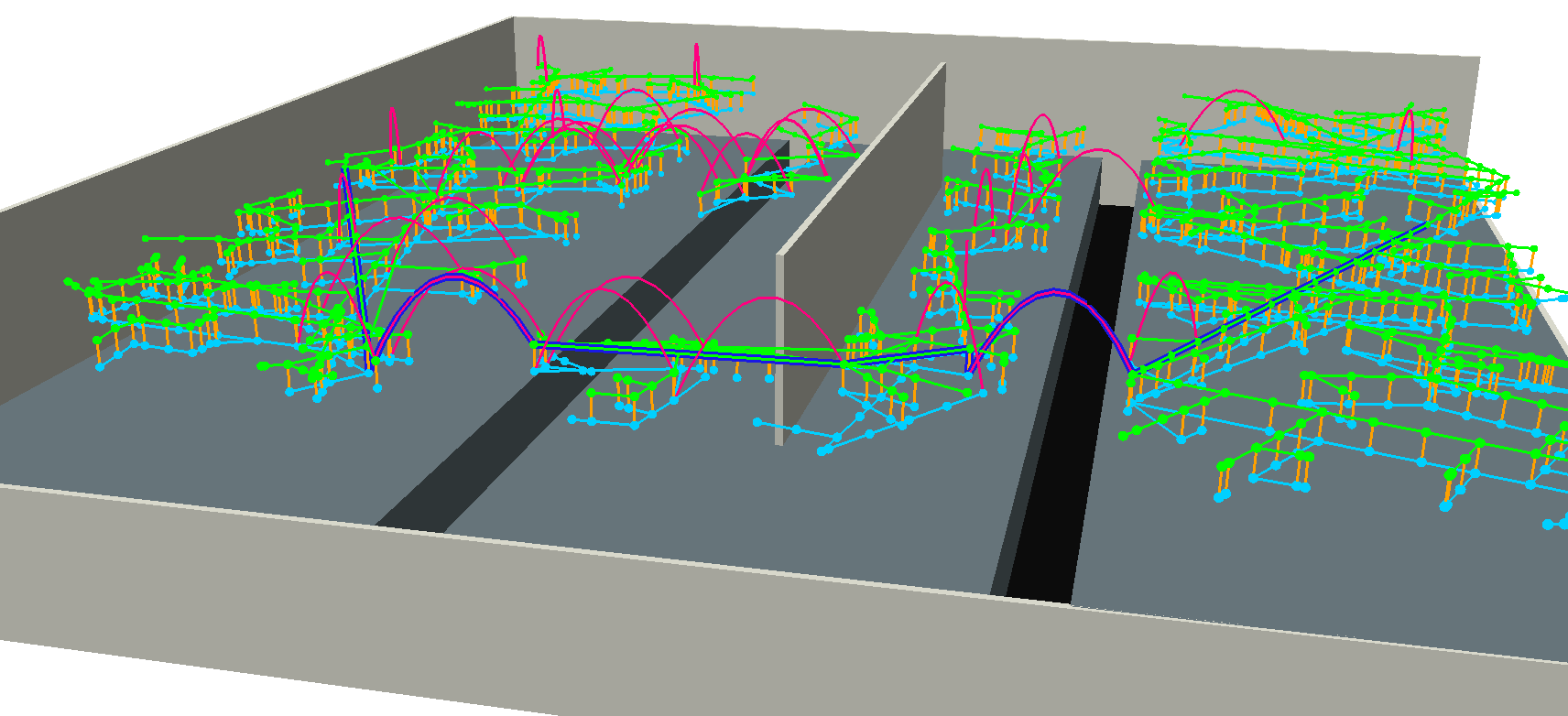}
  \end{subfigure}
  
  \caption{\label{fig:doublejump} The ``double jump'' scenario. The robot must jump across two gaps and navigate around a wall in the middle to get from the right side to the left side.}
\end{figure}
\begin{figure}
  \centering
  \begin{subfigure}[b]{0.75\linewidth}
    \captionsetup{justification=centering}
    \centering
    \includegraphics[width=\linewidth]{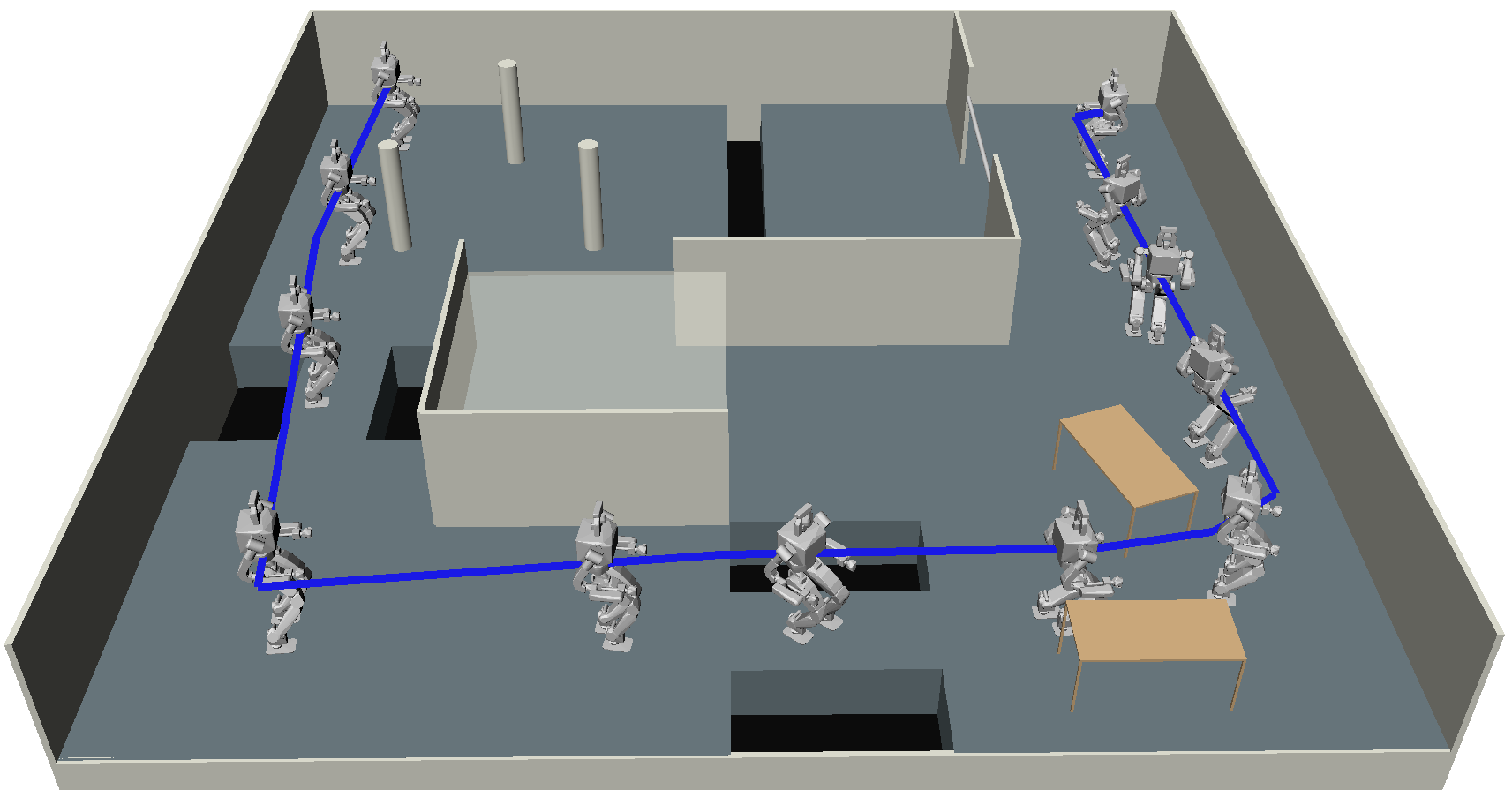}
    \caption{}
  \end{subfigure}
  
  \begin{subfigure}[b]{0.75\linewidth}
    \captionsetup{justification=centering}
    \centering
    \includegraphics[width=\linewidth]{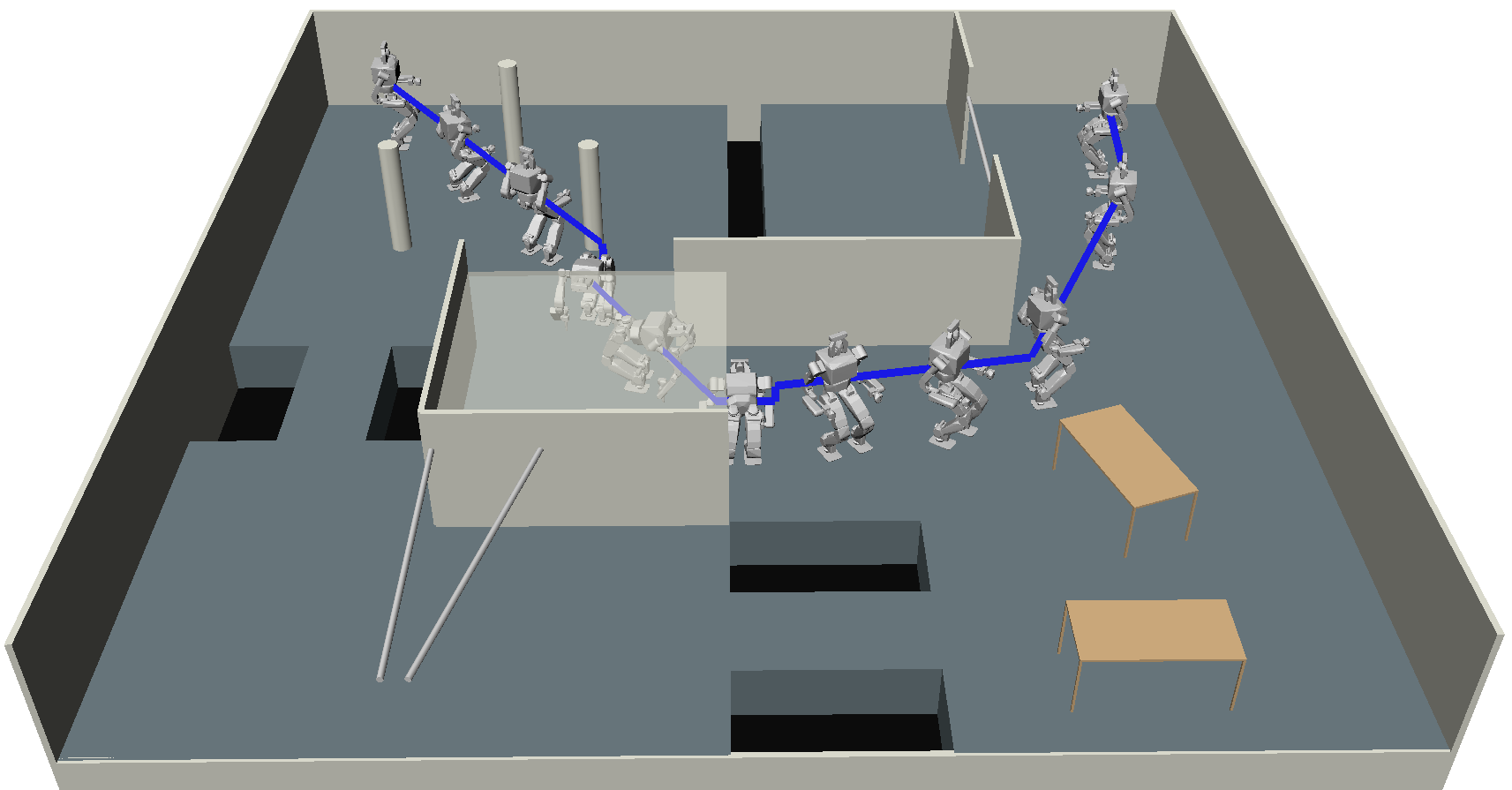}
    \caption{}
  \end{subfigure}
  
  \begin{subfigure}[b]{0.75\linewidth}
    \captionsetup{justification=centering}
    \centering
    \includegraphics[width=\linewidth]{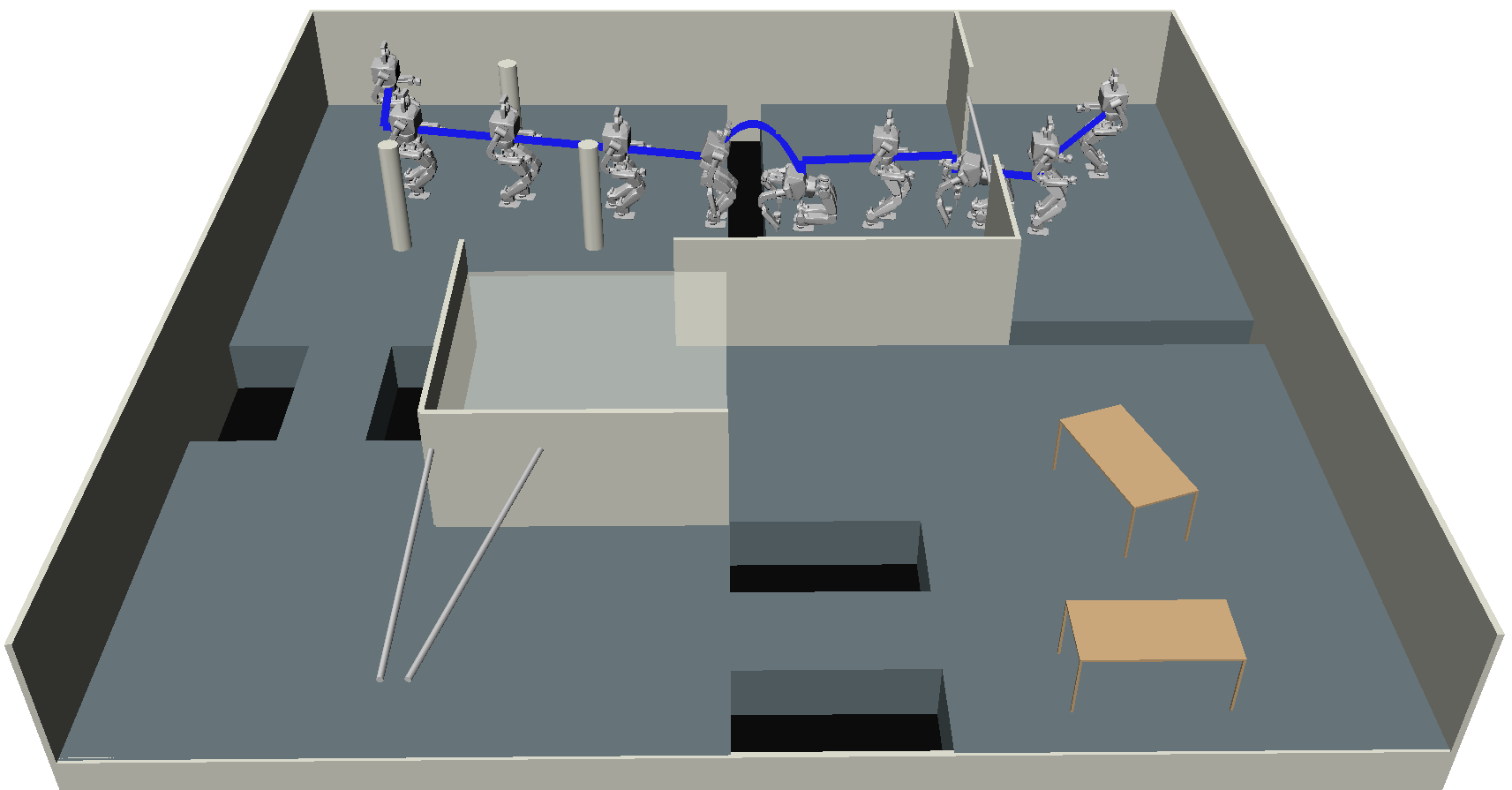}
    \caption{}
  \end{subfigure}
  
  \begin{subfigure}[b]{0.75\linewidth}
    \captionsetup{justification=centering}
    \centering
    \includegraphics[width=\linewidth]{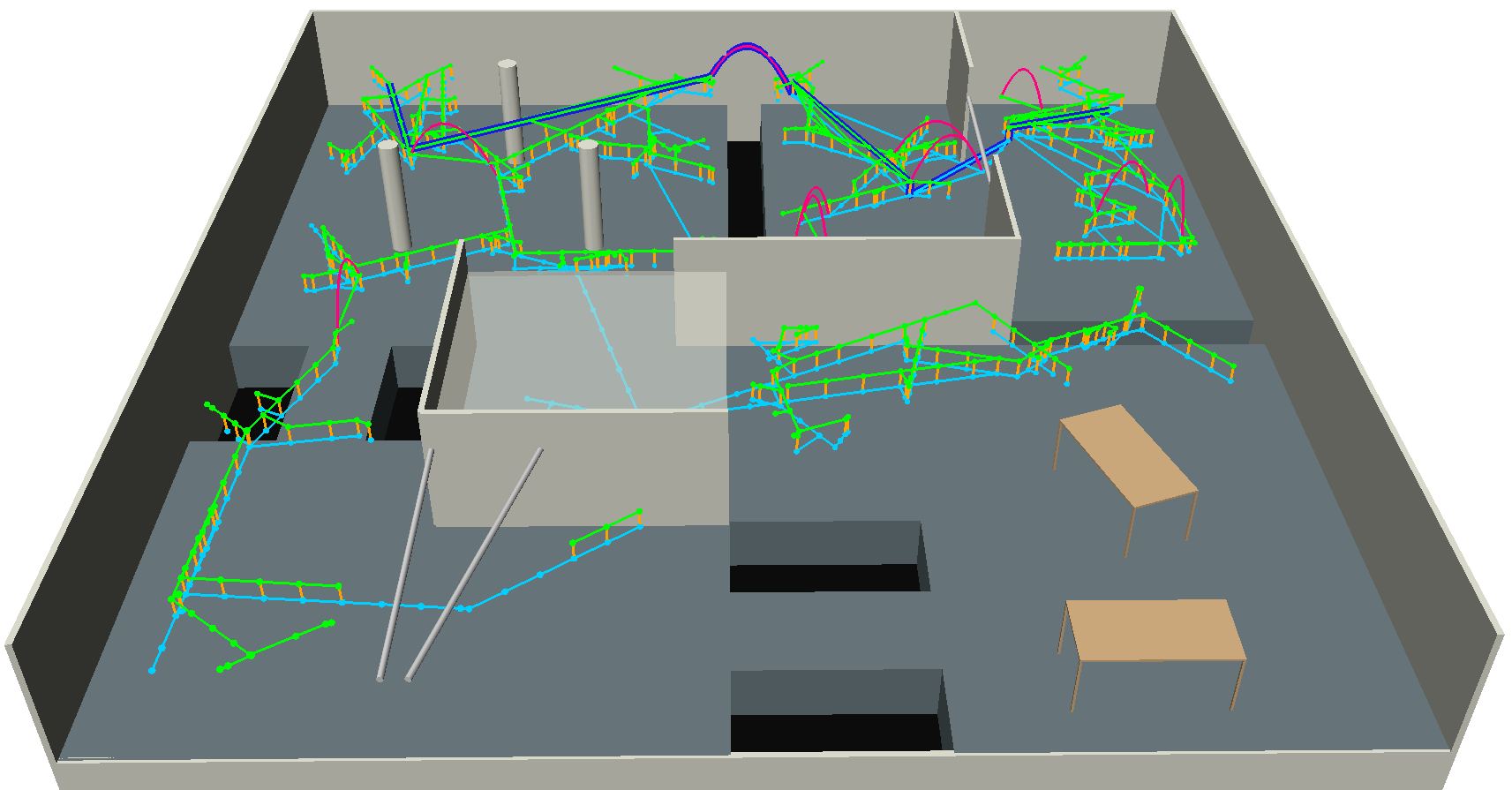}
    \caption{}
  \end{subfigure}
  
  
  
  
  \caption{\label{fig:threeroutes} The three versions of the ``Three routes'' scenario. The robot must get from the back left corner to the back right corner. (a) A route exists that allows the robot to walk all the way to the goal. (b) Some bars were added to the walking route, so the robot must crawl at least once to reach the goal. (c) A gap was added at the end of the crawling routes, so the robot must jump at least once to reach the goal. (d) A grid that shows the map being explored.}
\end{figure}

\begin{itemize}
  \item[] \textbf{Three Routes} scenario is shown in Fig. \ref{fig:threeroutes}. There are three potential routes that the robot might take to get from the start to the goal. We have three different versions of this scenario, and each version has progressively stronger requirements for what actions are needed by the solution, allowing us to compare the performance impact caused by specific action sequences being required.
  \item[] \textbf{Hallway} scenario was shown in Fig. \ref{fig:hallway}. The robot must crawl underneath some bars and then jump across a gap to get from the start on the right side to the goal on the left.
  \item[] \textbf{Double Jump} scenario is shown in Fig. \ref{fig:doublejump}. The robot must jump twice to get from the right side to the left.  
\end{itemize}

In Table \ref{tab:results} we see that the time required to solve a problem scales up with the number of actions being used (comparing the values in the \textbf{Graph} column of rows 1--3 or of 4--5). For every action that is utilized by the planner, more exploration needs to be performed, which tends to increase the runtime. Not only does the action's space get explored, but also the transitions between the actions need to be explored. However, this cost is additive, not multiplicative, so the overall costliness will be related to the sum (not product) of the costliness of the individual actions.


We can also see that the time required to solve a problem scales up with the number of actions \emph{required} by the environment to get a solution (comparing the \textbf{Graph} values of row 2 to 4 or of row 3 to 5). This is not surprising since requiring certain actions can be viewed as tightening the constraints on the solution, and tighter constraints tend to take longer to solve with randomized search.

\begin{table}
\caption{\label{tab:results}Time performance results, tested on an Intel\textsuperscript{\textregistered} Xeon\textsuperscript{\textregistered} Processor E3-1290 v2 (8M Cache, 3.70 GHz) with 16GB of RAM. $N_a$ is the number of action types that were provided to the planner. ``Graph'' is the time it took to generate a solved graph. ``Motion'' is the time it took to generate the physical motions for the solution. ``Success Rate'' is how many times the planner succeeded (instead of timing out). All times are given in seconds. Each result is the average of 50 runs; the standard deviation is given in parentheses.}
\begin{tabular}{|l|c||l|l|r|}
\hline
\textbf{Scenario} & \textbf{$N_a$} & \textbf{Graph} & \textbf{Motion} & \textbf{Success}\\\hline
Three Routes (a) & 1 & 0.088 (0.048) & 8.47 (0.81) & 100\%\\\hline
Three Routes (a) & 2 & 0.134 (0.076) & 8.75 (0.91) & 100\%\\\hline
Three Routes (a) & 3 & 0.484 (0.450) & 7.52 (1.86) & 100\%\\\hline
Three Routes (b) & 2 & 0.152 (0.112) & 9.23 (1.09) & 100\%\\\hline
Three Routes (b) & 3 & 0.561 (0.502) & 7.59 (2.30) & 100\%\\\hline
Three Routes (c) & 3 & 1.210 (0.218) & 5.73 (1.79) & 100\%\\\hline
Hallway          & 3 & 3.67 (11.52)  & 8.29 (0.84) & 96\%\\\hline
Double Jump      & 3 & 1.48 (0.34)   & 4.32 (0.28) & 100\%\\\hline
\end{tabular}
\end{table}


\section{Conclusions}\label{sec:conclusion}

We presented performance results of multi-action traversal plans being generated for the DRC-HUBO1 platform in complex environments. The complexity of the environments is derived from the fact that they require a variety of different action types to be interlaced in the correct sequence in order to navigate from the start to the goal. Three action types were used to traverse these environments: walking, crawling, and the standing long jump. The time required to fully generate the motion plans was less than 1/100th of the time that the motions require for physical execution. This makes the Possibility Graph a promising option for online use. Moreover, the time required to guarantee that a solution exists is even smaller, which suggests that the Possibility Graph would be an effective tool for higher-level task planners such as the Hybrid Backward-Forward planner \cite{garrett2015backward, grey2016hbf} which only needs to know whether a query is solvable.

The theoretical framework of the Possibility Graph can extend beyond the applications seen here. Future work will incorporate highly dynamic actions, e.g. running jumps, using nonlinear constrained optimization. This will open the door to fast, global, dynamic planning for high dimensional systems.

\newpage

\section*{Acknowledgments}
This work was supported by DARPA grant D15AP00006.

\bibliographystyle{IEEEtran}
\bibliography{icra2017.bib}

\end{document}